%% file: twopoint_calib.tex
\documentclass[10pt,twocolumn,letterpaper]{article}

\usepackage{wacv}
\usepackage{times}
\usepackage{epsfig}
\usepackage{graphicx}
\usepackage{amsmath}
\usepackage{amssymb}
\usepackage{booktabs}
\usepackage{color}
\usepackage{multirow}
\usepackage{url}

\include{macro} 

\wacvfinalcopy 

\ifwacvfinal\fi
\begin{document}

\title{A Two-point Method for PTZ Camera Calibration in Sports}

\author{Jianhui Chen$^{*}$ \qquad Fangrui Zhu$^\dagger$ \qquad James J. Little $^*$\\
$^*$University of British Columbia \qquad $^\dagger$Tongji University\\
{\tt\small \{jhchen14,fangru1,little\}@cs.ubc.ca}
\and
}

\maketitle
\ifwacvfinal\thispagestyle{empty}\fi

\footnotetext[1]{{This work was supported by grants from NSERC. The majority of work was performed when Fangrui Zhu visited UBC as a Mitacs intern.}}
\begin{abstract}
Calibrating narrow field of view soccer cameras is challenging because there are very few field markings in the image. Unlike previous solutions, we propose a two-point method, which requires only two point correspondences given the prior knowledge of base location and orientation of a pan-tilt-zoom (PTZ) camera. We deploy this new calibration method to annotate pan-tilt-zoom data from soccer videos. The collected data are used as references for new images. We also propose a fast random forest method to predict pan-tilt angles without image-to-image feature matching, leading to an efficient calibration method for new images. We demonstrate our system on synthetic data and two real soccer datasets. Our two-point approach achieves superior performance over the state-of-the-art method.       
\end{abstract}


\section{Introduction}
Computer vision for sports analysis has the potential to increase a team's competitive ability, identify talented prospects in junior leagues and give fans the statistics of their favorite players. Moreover, companies like SPORTLOGiQ \cite{sportslogic} extract player statistics from widely available broadcasting videos. At the same time, broadcasting companies like ESPN are building automatic broadcasting systems \cite{chen2016learning,chen2017should} that also require labeled data from the already existed broadcasting videos.

The sports analysis and autonomous camera systems (ACSs) require robust camera parameter estimation. First, most statistics (\eg player trajectory and team format) are measured in model (\eg  playing field of soccer) coordinates. Hence, to view the data in model coordinates requires the registration of the broadcasting viewpoint with the model. Second, ACSs require a large amount of camera parameter data as labels to learn a stable broadcast system from human operators. In practice, human operators capture the game from various types of views such as top views and side views. To build an autonomous camera system for different views, calibration methods are in high demand for collecting ground truth data from these views. Our work is motivated by these real applications.

\begin{figure}
	\begin{center}
		\includegraphics[width = 1.0\linewidth]{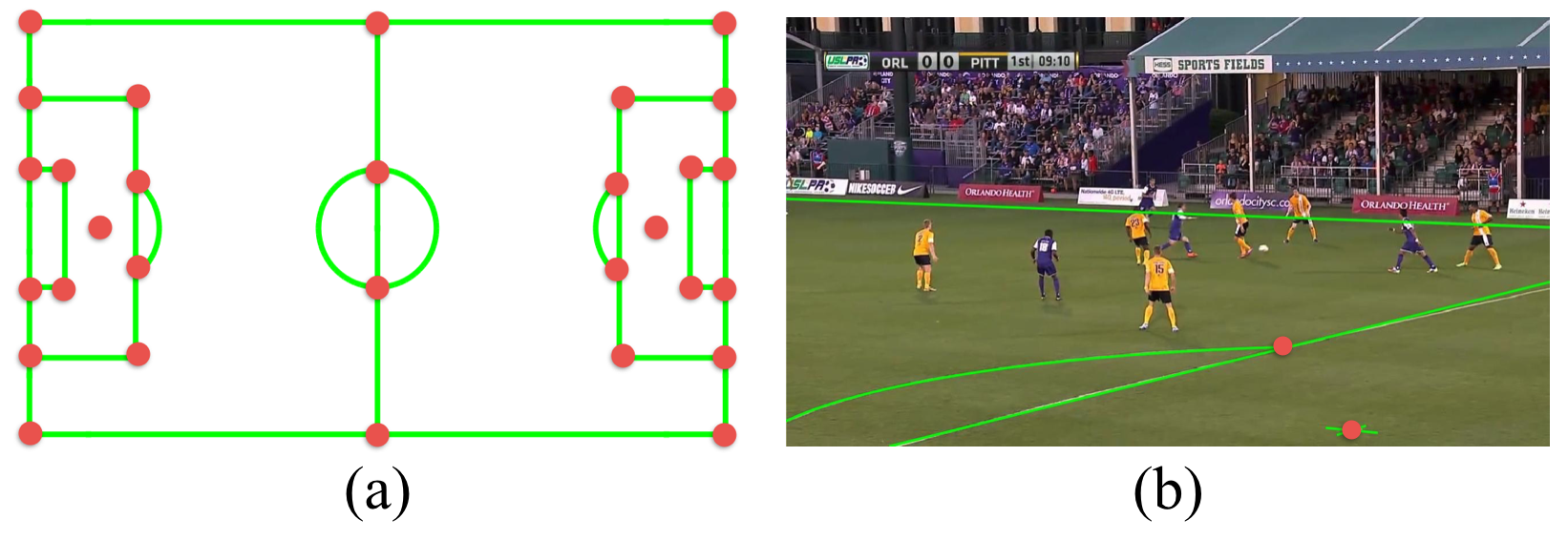}  
	\end{center}
	\vspace{-4mm}
	\caption{(a) Soccer field model. The red points are marking intersections and two penalty marks. These points are generally used in human annotation. (b) A narrow field of view image with overlaid markings (green lines). There are only two observable model points (red circles) in the image, making minimum four point based correspondences annotation difficult. In this work, we propose a two-point algorithm to calibrate cameras of images like (b).}
	\label{fig:model_line}
\end{figure}

Camera calibration for soccer games has been studied extensively in the last two decades \cite{hayet2004robust,puwein2012ptz,aleman2014camera,alvarez2016real}. The problem can be formulated as a homography estimation problem. Estimating the homography requires at least 8 constraints which can be 4 point-to-point correspondences or 3 point-to-point plus 2 point-on-line correspondences \etc. The requirement of 8 constraints may not be satisfied when cameras look at a small area of the model, as in Figure~\ref{fig:model_line}(b). Because a considerable percentage of images have fewer constraints (\eg narrow field-of-view cameras), less-constrained algorithms are needed.

Most previous approaches are based on reference images annotation and points matching \cite{hayet2004robust,hess2007improved,schonberger2017comparative}. At first, the camera parameters of the reference image are estimated from human annotated correspondences with satisfactory accuracy. The references  are stored in a database with corresponding camera parameters and keypoint features. Then, a new image queries the database to find the most similar reference image. The camera of the new image is calibrated by matching keypoints (\eg SIFT \cite{lowe2004distinctive}) from a reference image and the new image. These approaches have been extended to automatic reference generation \cite{wen2016court,homayounfar2017sports} and line/ellipse matching \cite{farin2005fast,gupta2011using,liu2017new}. However, human annotation is still needed to get ground truth data and a small set of reference images. Moreover, to the best of our knowledge, estimating narrow FOV PTZ cameras for soccer cameras is still an open problem in either manual annotation or automatic calibration \cite{sinha2006pan,del2010exploiting,wu2013keeping,lisanti2016continuous}. 

In this work, we propose a two-point method to calibrate soccer game cameras. In soccer broadcasting, the PTZ camera does not change its location and base rotation under some tolerance. This phenomenon has been observed by Thomas \cite{thomas2007real} and has been used to estimate PTZ configurations. Our work advances this research's direction to use the PTZ base information as a prior in camera data annotation and automatic camera calibration. With the prior, our method requires only two points to calibrate a PTZ camera. Figure \ref{fig:model_line} shows a soccer field model and an example image calibrated by our method.

When the FOV is narrow, calibrating new images automatically using conventional reference image methods is time consuming. Because each reference image only covers a small part of the environment (\eg stadium buildings and commercial boards around the field), the new image query has to search through a large number of references to find the most similar one which still does not always provide sufficient matching points. To address this problem, we propose a \textit{pan-tilt forest} algorithm to effectively associate the data from many reference images. During testing, the pan-tilt forest directly predicts the pan-tilt angles from local patch descriptors without image-to-image feature matching, greatly speeding up the algorithm. In summary, our paper has two main contributions:
\begin{itemize}
\setlength{\itemsep}{2pt}	
  	\item a two-point algorithm to calibrate a fixed location PTZ camera,
  	\item a pan-tilt forest algorithm to robustly predict pan/tilt angles given local patch descriptors and speed up the automatic PTZ camera calibration.
\end{itemize}

In the following, we start with a discussion of related work, and then describe our method. Finally, we apply our method to a synthetic dataset and two real soccer datasets. The experiments show that our method is robust to noise from feature point locations and camera locations. In addition, our method successfully calibrates images on both real datasets with higher accuracy than one recently developed method, and it requires less running time, demonstrating its practical value in real applications. 

\section{Related Work}
Pan-tilt-zoom (PTZ) cameras have been used in surveillance and events broadcasting~\cite{wheeler2010face,park2013face,neves2015acquiring,homayounfar2017sports}. Point and line feature matching is widely used to estimate the homography matrix between the playing field model and the image \cite{zeng2008new,dubrofsky2008combining}. Point features have been used to estimate the camera parameters \cite{lowe2004distinctive,hess2007improved, meng2016exploiting}. Okuma \etal \cite{okuma2004automatic} and Thomas \cite{thomas2007real} have proposed line feature based methods which are more robust under circumstances when point features are not descriptive such as in ice hockey and soccer games. Other feature based methods such as \cite{ghanem2012robust} have been proposed in consequence of insufficient visual information provided by point/line features.

Recently, Homayounfar \etal \cite{homayounfar2017sports} has proposed a fully automatic sports field relocalization method by formulating the problem as a branch and bound inference in a Markov random field where an energy function is defined in terms of semantic cues (\eg field surface, lines and circles) obtained from a deep semantic segmentation network. Their method achieves high accuracy which is measured by IoU (Intersection over Union) between ground truth and predicted playing field areas in both soccer and ice hockey games.   

Moreover, Sharma \etal \cite{sharma2017automated} has proposed a method to formulate the calibration as a nearest neighbour search problem over a synthetically generated dictionary of edge map taking from line segments and homography pairs. This approach tackles the problem of insufficient suitable point correspondences in the case of soccer fields. However, it is mainly applied to top zoom-out views.

\section{Method}

\begin{figure}
	\begin{center}
		\includegraphics[width = 1.0\linewidth]{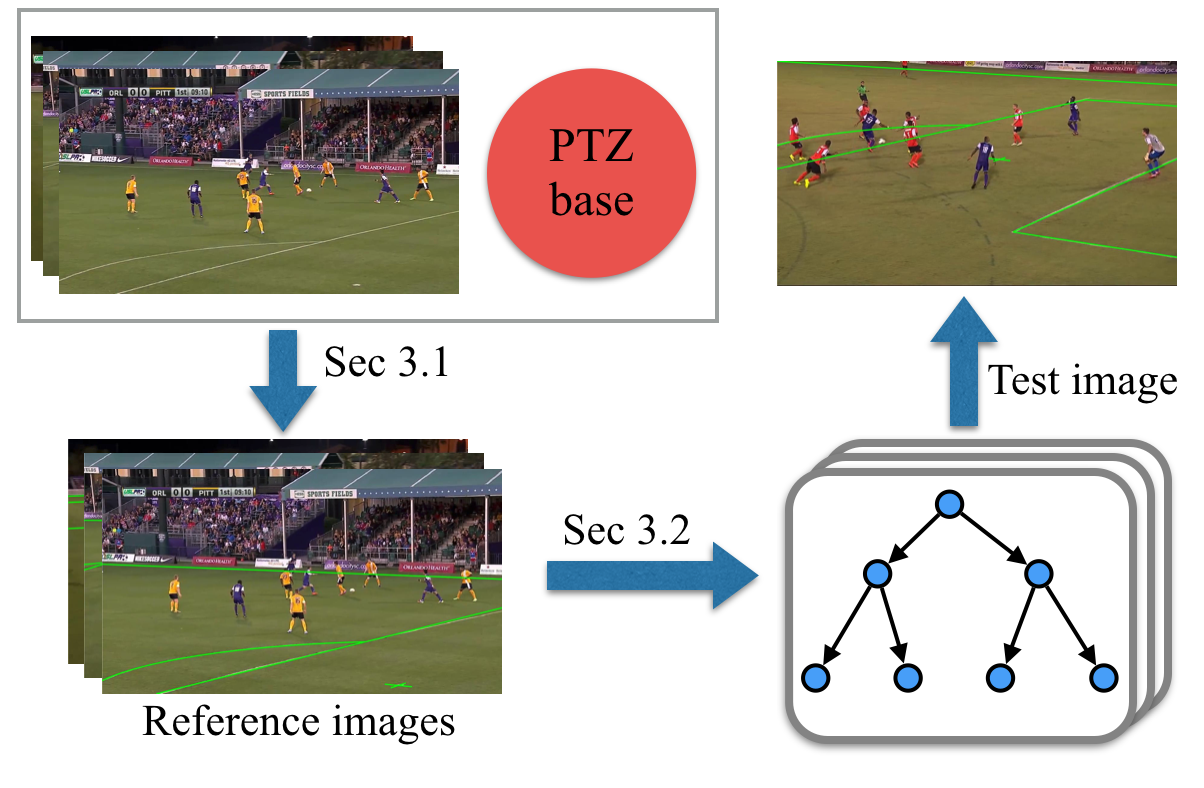}  
	\end{center}
	\vspace{-4mm}
	\caption{\textbf{Pipeline of our system.} Given the PTZ camera base information, we first manually annotate reference images using a two-point algorithm. We then use the pan-tilt forest method to encode the reference images. In testing, the predictions from the pan-tilt forest are used to calibrate a new image.}
	\label{fig:pipe_line}
\end{figure}

We use the pinhole model to describe a PTZ camera
\begin{equation}
	\mathtt{P} = 
	\underbrace{\mathtt{K} \mathtt{Q}_{\phi}\mathtt{Q}_{\theta}}_\text{PTZ} 
	\underbrace{\mathtt{S} [\mathtt{I} | -\mathbf{C}]}_\text{prior},
	\label{equ:pinhole}
\end{equation}
where $\mathbf{C}$ is the camera's center of projection. The combination of $\mathtt{Q}_{\phi}\mathtt{Q}_{\theta} \mathtt{S}$ describes rotations from world to camera coordinates. First, it rotates the camera to the PTZ camera base by $\mathtt{S}$. Then the camera pans by $\mathtt{Q}_{\theta}$ and tilts by $\mathtt{Q}_{\phi}$. $\mathtt{K}$ is the intrinsic matrix. Assuming square pixels and a principal point at the image center $(u, v)$, the focal length $f$ is the only unknown variable in $\mathtt{K}$.

$\mathtt{P}$ can be separated into two parts. The right part $\mathtt{S} [\mathtt{I} | -\mathbf{C}]$ is time invariant for a PTZ camera with a fixed location. This part can be estimated using images from various PTZ configurations \cite{thomas2007real} and used as a prior. 

Because the camera only pans, tilts and zooms, the left part $\mathtt{K} \mathtt{Q}_{\phi}\mathtt{Q}_{\theta}$ projects a ray $\br = (\theta_\p, \phi_\p)$ to the image location $\p = (x, y)$:
\begin{equation}
\label{equ:project_ray}
\p = \mathtt{P}(\br) = [f\tan(\theta_\p - \theta) + u, f\tan(\phi_\p - \phi) + v]^T,
\end{equation}
where we parameterize the ray by pan/tilt angles of its pixel location $\p$. Specially, when $\br = (\theta, \phi)$, the ray passes the image center. Note that the pan/tilt angles of a particular camera can be estimated by two pairs of $\{\br, \p\}$ using \eqref{equ:project_ray}. This observation is the initial motivation for the present work.

Figure \ref{fig:pipe_line} shows the pipeline of our method. We introduce a two-point algorithm to manually annotate reference images. Then, we introduce the pan-tilt forest algorithm to automatically calibrate new images.

\subsection{Two-point Algorithm for Data Annotation}
\label{sec:two_point_calib}
Given two 3D-2D point correspondences (\eg from human annotation), our method first estimates the focal length from two points \cite{gedikli2009continual}. Then it estimates the initial pan/tilt angles using one point \cite{li2015method}. Finally, it optimizes the PTZ parameter by minimizing the reprojection error using both points. The components of the method have been proposed by Gedikli \etal \cite{gedikli2009continual} and Li \etal \cite{li2015method}, separately. However, to the best of our knowledge, we are the first to integrate them and apply them to sports camera calibration.

\paragraph{Estimate Focal Length from Two Points}
Two 3D points $\X_1$ and $\X_2$ in the world coordinate are projected to image points $\x_1$ and $\x_2$ respectively. $\x_1$ and $\x_2$ can be back-projected to two rays $\textbf{d}_1$ and $\textbf{d}_2$ in the camera coordinate. The cosine formula for the angle $\alpha$ between two rays is:

\begin{equation} 
\label{equ:angle}
\begin{split}
\cos{\alpha} & = \frac{\bd_1^T\bd_2}{\sqrt[]{\bd_1^T\bd_1}\sqrt[]{\bd_2^T\bd_2}} \\
			 & = \frac{({\tK^{-1}\x_1})^T({\tK^{-1}\x_2})}{\sqrt[]{({\tK^{-1}\x_1})^T (\tK^{-1}\x_1) }\sqrt[]{({\tK^{-1}\x_2})^T ({\tK^{-1}\x_2}})} \\
 			 & = \frac{\x_1^T\omega \x_2}{\sqrt[]{\x_1^T\omega \x_1} \sqrt[]{\x_2^T\omega \x_2}},
\end{split}
\end{equation}
where $\omega$ is the image of the absolute conic (IAC) and it is related to the intrinsic matrix by $\omega = {\tK}^{-T}{\tK}^{-1}$. The angle $\alpha$ is determined by three known points $\mathbf{C}$,  $\X_1$ and $\X_2$. Thus the focal length $f$ is the only unknown element in \eqref{equ:angle}. It can be solved in the standard quadratic equation \cite{gedikli2009continual} ( details in Appendix \ref{subseq:app_focal_length}).

\paragraph{Estimate Pan, Tilt from One Point}
With known focal length, a PTZ camera becomes a pan, tilt (PT) camera. Li \etal \cite{li2015method} has proposed a method for PT camera calibration using a single point. The method first builds a nonlinear PT camera model with respect to pan and tilt angles. Then a closed-form solution of pan and tilt can be derived by solving a quadratic equation of tangent of pan. The solution for our camera model is in Appendix \ref{subsec:app_single}.

With initial values of pan, tilt and focal length, our method further optimizes parameters by minimizing the reprojection error using Levenberg-Marquardt optimization. Figure \ref{fig:model_line}(b) shows an example of the two-point method. There are fewer than 4 observable points (i.e.\ marking intersections and penalty marks) in the image. In that situation, the classic four-point method is not applicable and our two-point method is very useful. 

\subsection{Pan-tilt Forests}
\label{sec:pan_tilt_forests}
We use a learning-based method to predict the ray $\br$ given its appearance in the image. We model this problem as a regression problem:
\begin{equation}
\hat{\br}_{\mathbf{p}} = h (\x_\mathbf{p}),
\end{equation}
\normalsize
where $\x_\mathbf{p}$ is the feature (\eg SIFT descriptor) that describes the image patch whose center is at $\mathbf{p}$. The label $\br_\p$ is given by:
\begin{equation}
\br_\p = [\theta + \arctan{\frac{x-u}{f}}, \phi+ \arctan{\frac{y-v}{f}}]^T,
\end{equation}
where ${\theta, \phi}$ are pan and tilt angles of the camera. In training, $\{(\x_\mathbf{p}, \br)\}$ are paired training data. In testing, the ray $\hat{\br}_{\mathbf{p}}$ is predicted by the learned regressor $h(\cdot)$. 

We use a random forest to predict the ray as it has achieved state-of-the-art in indoor and outdoor camera pose estimation \cite{cavallari2017onthefly,meng2017backtracking} and denote it as \textit{pan-tilt forest}. 

A random forest is an ensemble of randomized decision trees. Each decision tree is a binary-tree consisting of split nodes and prediction nodes. Following a top-down approach, the decision tree optimizes parameters $\pi_n$ of $n^{th}$ node from the root node and recursively processes the child nodes till the leaf nodes. The parameter $\pi_n$ is chosen from a set of randomly sampled candidates $\Pi_n$. 

At each split node $n$, decision trees learn the $\pi_n$ that \doubleQuote{best} splits the incoming training set $S_n$ into $S_n^L$ and $S_n^R$ which are training sets of left and right sub-trees, respectively. This problem is formulated as the maximization of the information gain at the $n^{th}$ node
\begin{equation}
\arg \max_{\pi_n \in \Pi_n} E(S_n, \theta),
\end{equation}
where
\begin{equation}
E(S_n, \theta) = \sum_{(\x, \br) \in S_n}(\br - \bar{\br})^2 - \sum_{j \in {L,R}}(\sum_{(\x, \br) \in S_n^j}(\br - \bar{\br})^2),
\end{equation}
where $\bar{\br}$ indicates the mean value of $\br$ for all training samples reaching the node. Note that left and right subsets $S_n^j$ are implicitly conditioned on the parameter $\pi_n$. Here we omit the subscript $\p$ of $\br_\p$ and $\x_\p$ for notation convenience.    

Training terminates when a node reaches a maximum depth $D$ or contains too few examples. In a leaf node, we store mean values of samples that reaches that leaf node $(\bar{\x}_{l}, \bar{\br}_{l})$. During testing, a sample traverses the tree $t$ from the root node to a leaf node. The leaf outputs the $\bar{\br}_{l}$ as the prediction. We also use the feature distance $\|\bar{\x}_{l} - \x \|^2_2$ (lower than a threshold) to remove outliers (\eg features on players). As we will show in the experiment, this pre-processing is very effective in practice. We keep predictions from all trees and choose the one that minimizes the reprojection error in the camera pose optimization process. 

\subsection{Camera Pose Optimization}
\label{sec:pose_opt}
From the pan-tilt forest, we obtain a set of pixel location and predicted ray pairs $\{(\p, \hat{\br})\}$. The camera pose optimization is to minimize the re-projection error:
\begin{equation}
\{\theta, \phi, f\} = \arg \min_{\mathtt{P}} \sum_{i}{\|\p_i - \mathtt{P}(\hat{\br}_i) \|^2}.
\end{equation}
Because the predictions from the pan-tilt forest may still have large errors, we use the RANSAC method \cite{fischler1981random} to remove outliers. In the internal loop of RANSAC, the minimal number of points required to fit the model is two. As RANSAC is a standard method to perform model estimation with the presence of outliers, the fewer iterations needed, the better. Table \ref{tab:RANSAC_iter_num} shows the number of iterations needed as a function of the minimum sets of points \cite{scaramuzza20111}. The assumptions are: probability of success=99\%, percentage of outliers=50\%. Our two-point method is better than the four-point based homography estimation method. That enables our method to find the optimal solution efficiently given the large ratio of outliers. 

\begin{table}[htbp]
    \centering
    \begin{tabular}{c|ccc}
    \hline
         Min. set of points &4 &2 &1 \\
         \hline
         No. of iterations &71 &16 &7\\
    \hline     
    \end{tabular}
    \vspace{4mm}
    \caption{\textbf{Number of RANSAC iterations.}}
    \label{tab:RANSAC_iter_num}
\end{table}

\section{Experiments}
\paragraph{Highlights Dataset}
This dataset was collected from the public soccer highlights dataset \cite{keyu17light}. It contains four image sequences from two professional soccer games (two sequences for each game). The image resolution is $1280 \times 720$ and the sample rate is about 6 FPS. The total frame number is 116.

The ground truth camera parameters are manually calibrated. We first calibrate the images that have at least four correspondences. Then we run a global optimization algorithm to estimate the PTZ camera base parameters. Then we calibrate the rest of the images using the two-point method. After that, we check camera parameters by projecting the model to images. This process is repeated until there are satisfactory results. 
\begin{figure}
    \begin{center}
        \includegraphics[width = 1.0\linewidth]{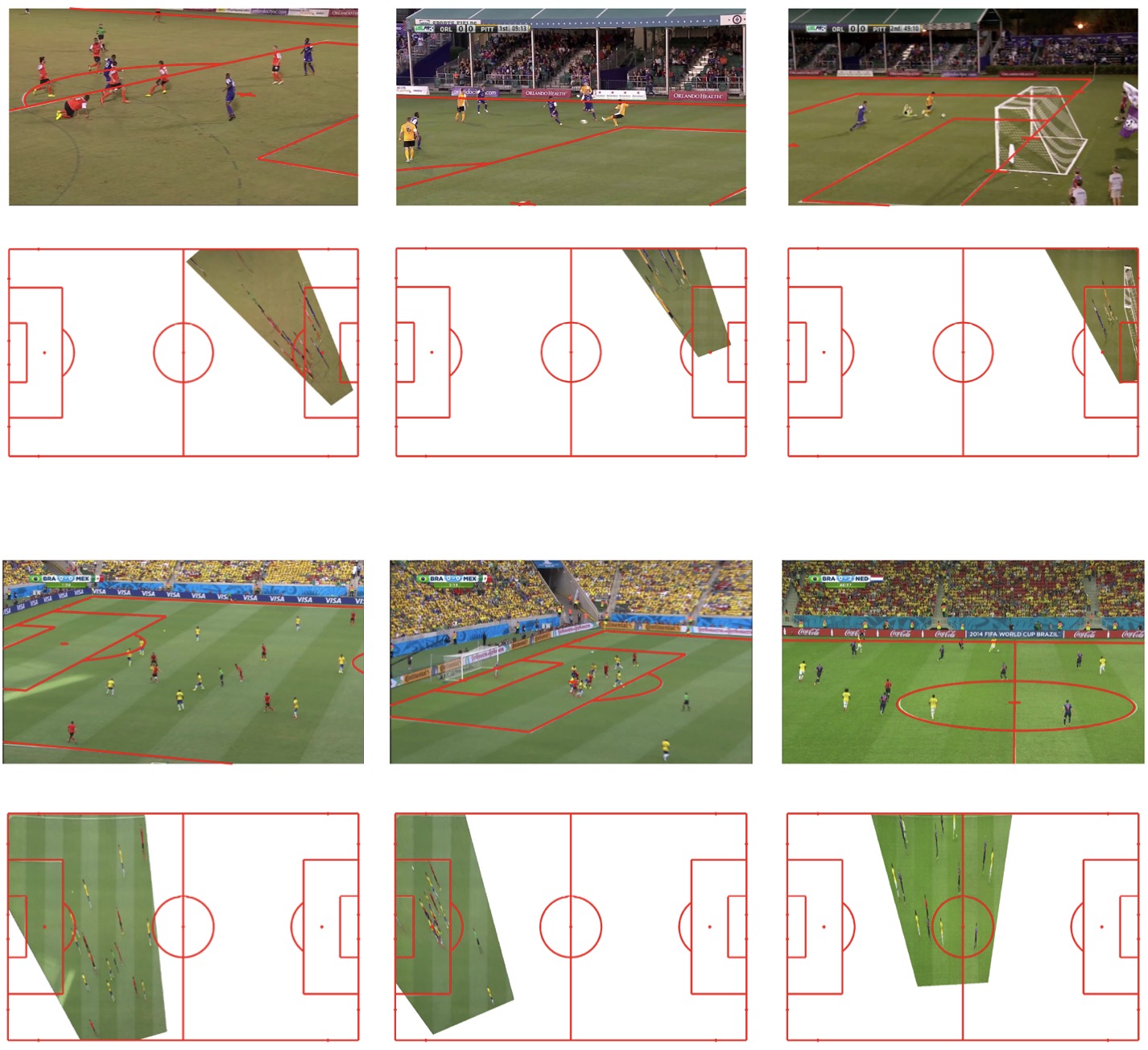}
    \end{center}
    \vspace{-4mm}
    \caption{\textbf{Examples of ground truth annotation.} Top row: highlights dataset; bottom row: World Cup dataset.}
    \label{fig:gt_annotation}
\end{figure}

\begin{figure}
    \begin{center}
        \includegraphics[width = \linewidth]{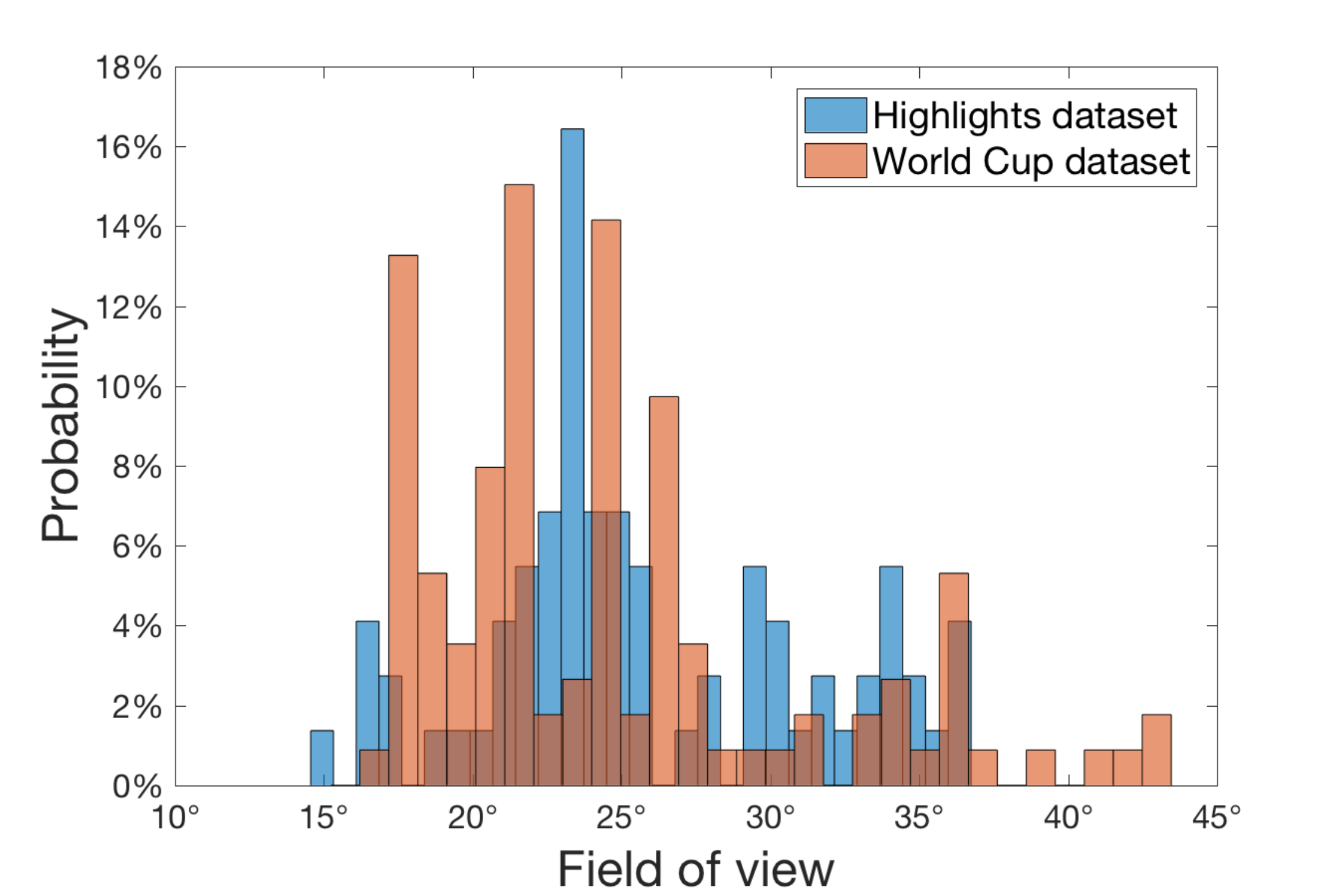}
    \end{center}
    \vspace{-2mm}
    \caption{\textbf{Distribution of field of view of the two datasets.}  }
    \label{fig:fov_distribution}
\end{figure}

\paragraph{World Cup Dataset}
This dataset was selected from the World Cup 2014 dataset collected by Homayounfar \etal \cite{homayounfar2017sports}. The original dataset has 20 games and has been used to automatically localize soccer fields. Because our method requires the prior knowledge of camera bases, we selected two games (BRA vs. MEX and BRA vs. NED) because they have sufficient number of images (42 and 33) to cover the whole stadium. The ground truth data was obtained by manual annotation. In these two games, we found that the assumption of fixed PTZ camera base generally holds in soccer broadcasting, providing evidence of potential wide applications of our method.

Figure \ref{fig:fov_distribution} shows the distribution of field-of-view of two datasets. Both datasets have narrow FOV images (smaller than $25^o$) although the World Cup dataset has some wide FOV images. Figure \ref{fig:gt_annotation} shows examples of ground truth annotation by overlapping the markings on the image and warping the image to the model. These example images demonstrate the correctness of our decomposition of the pinhole camera into \doubleQuote{prior} and \doubleQuote{PTZ} parts.

\subsection{Synthetic Data Experiments}

\begin{figure}
    \centering
    \includegraphics[width = 1.0\linewidth]{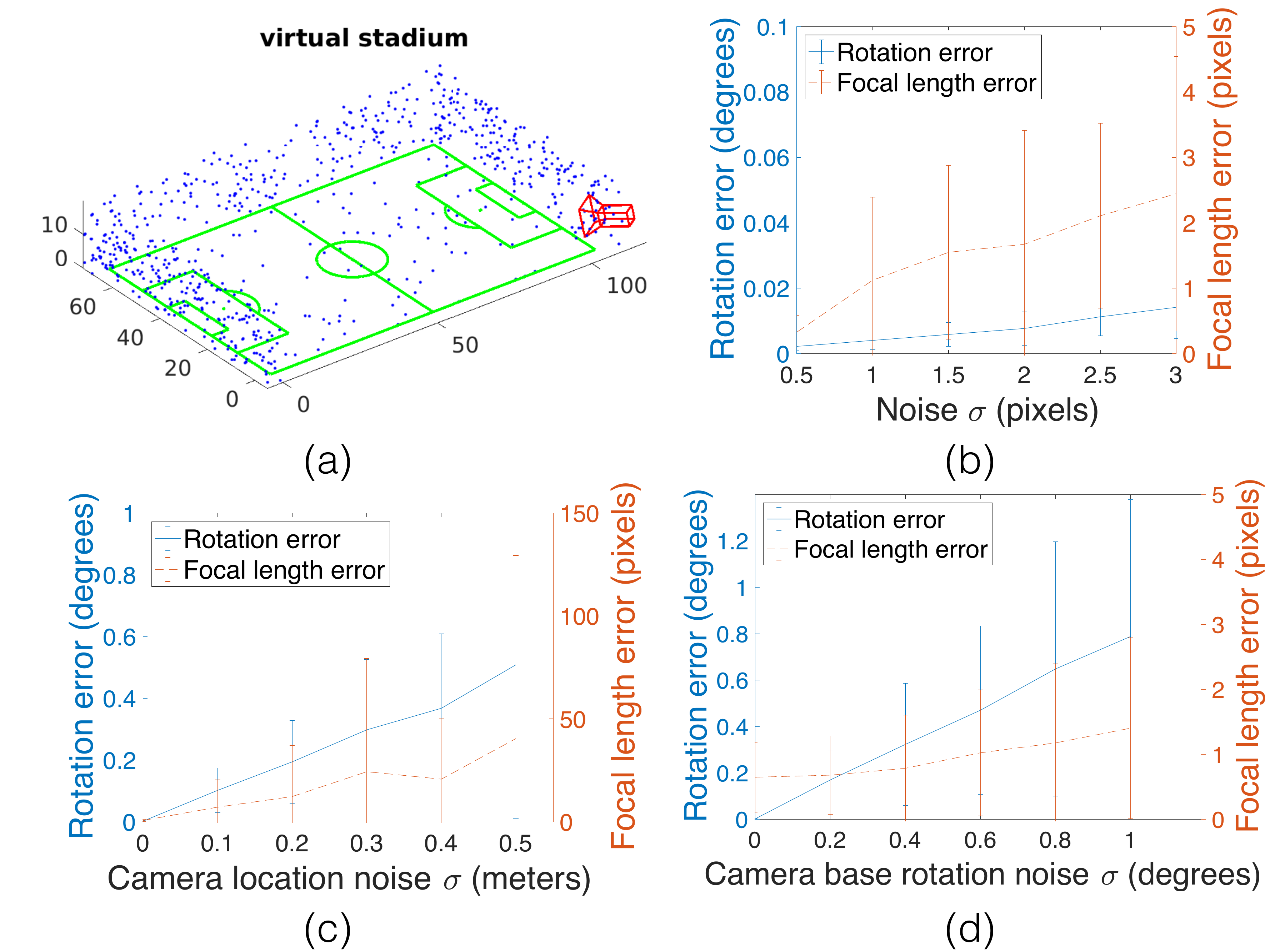}
    \caption{\textbf{Synthetic experiments results.} (a) An example of the synthetic stadium used in the experiment. (b-d) Mean errors as a function of the noise in feature location, camera base location and rotation, respectively. Error bars show standard deviations. }
    \label{fig:syn_result}
\end{figure}
To estimate the accuracy of the method under typical feature location noise conditions, we conducted experiments with synthetic data. We set the camera base parameter as the one we estimated from the highlights dataset. That camera is located on the right corner of the playing field. Then, we randomly set pan, tilt and focal length values in the range of $[15^o,75^o]$, $[-14^o, -5^o]$ and $[1500, 5000]$, respectively. Then, we randomly sample 200 rays in which about $90\%$ of them are projected out of the playing field. The number of $90\%$ is chosen experimentally to simulate keypoint distribution in real data. Finally, we add different levels of Gaussian noise to disturb the feature locations in the image and estimate the PTZ configuration using our method. The experiment was performed on 100 cameras. For each camera, we repeated the experiment 100 times with different random values. 

Figure \ref{fig:syn_result}(b) shows the mean errors of rotation and focal length as a function of noise levels. Up to $\sigma = 3.0$, the mean rotation error is smaller than $0.02^o$ and the mean focal length error is smaller than 2.5 pixels, demonstrating the stability of our method in terms of Gaussian noise in feature locations. 

Because the estimation of camera base location and orientation may have errors in practice, we study the influence of estimation errors on our method. Figure \ref{fig:syn_result}(c,d) show the PTZ parameter errors as a function of camera base location and rotation uncertainties, respectively. Our method is robust to the uncertainties from camera location (see Figure~\ref{fig:syn_result}(c)). One explanation is that the movement of camera location is relatively small compared to the spatial extent of the stadium. On the other hand, the rotation error of our method is sensitive to camera base rotation uncertainties. It is not unexpected as the base rotation error can be directly propagated from the \doubleQuote{prior} part to the \doubleQuote{PTZ} part in \eqref{equ:pinhole}. We report the absolute errors of the estimated focal length in Figure \ref{fig:syn_result}. The relative error of 100 pixels is about 3.2\% of the ground truth.

\subsection{Real Data Experiments}
\begin{figure*}
	\begin{center}
		\includegraphics[width = 0.95\linewidth]{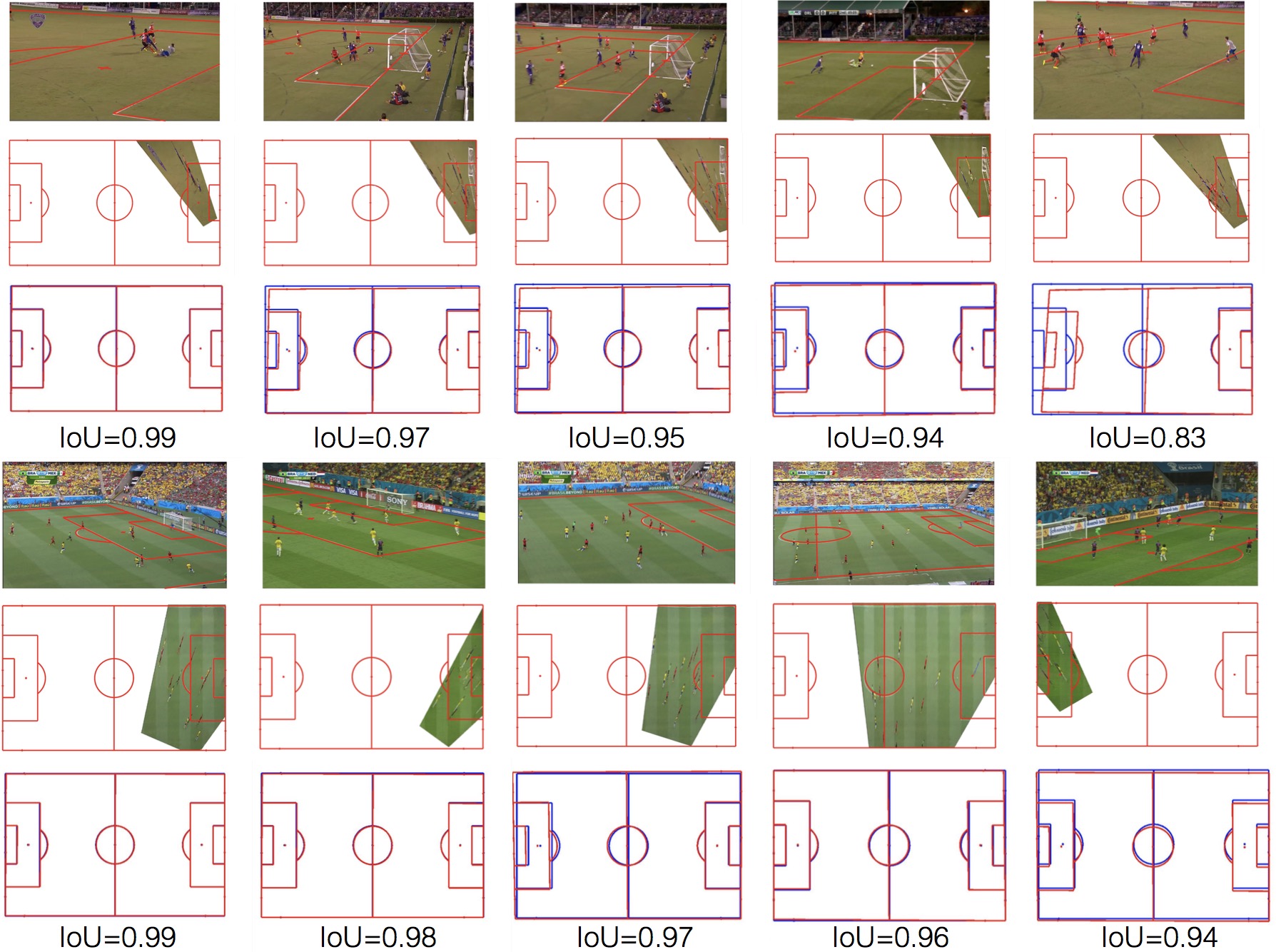}  
	\end{center}
	\vspace{-2mm}
	\caption{\textbf{Qualitative results and corresponding IoU scores.} Top row: results on highlights dataset; bottom row: results on World Cup dataset. In the top row, the most right column shows a typical challenging case for our method. The image has a very small texture-rich area, resulting in insufficient number of SIFT features for our method. }
	\label{fig:qualitative_result}
\end{figure*}

\paragraph{Error Metric}
We use the IoU (Intersection over Union) \cite{homayounfar2017sports} to measure the calibration accuracy of different methods. The IoU is calculated by warping the projected model to the topview by the ground truth camera and the estimated camera (see the third row in Figure \ref{fig:qualitative_result}). For our method, we also report the rotation and focal length error distribution.

In the highlights dataset, we employ leave-one-sequence-out cross validation (\eg train on sequence 1, 2, 3 and test on sequence 4) in the experiment. In the World Cup dataset, half of the images were held out for training and the rest were used for testing in each game.

We compare our method with the CalibMe method \cite{chen2017should}, a representative reference image based method that has achieved good performance on soccer videos. For each sequence, we manually select 3-5 reference images to cover the whole stadium to make sure the CalibMe method works well. The number of the references is limited by the running time budget. 

\paragraph{Main Results}
\begin{table*}
 \centering
 \scalebox{0.9}
 
  \begin{tabular}{c|ccc|cc}
    
    \hline
    
    \multirow{2}{*}{Method} &  \multicolumn{3}{|c|}{Range}  & \multirow{2}{*}{CalibMe \cite{chen2017should}} & \multirow{2}{*}{Ours}\\
       \cline{2-4}
      & Pan & Tilt & Focal length &  & \\
  
    \cline{1-6}
    
    Seq1 & [49,68] & [-13, -9] & [1586,4346] 
    &$0.75\pm0.22$ & $\mathbf{0.88}\pm0.06$\\

    Seq2 &[49,68] &[-11, -8] & [3330, 4947]  &$0.73\pm0.27$ & $\mathbf{0.81}\pm0.21$\\

    Seq3 & [17,67] & [-9,-5] & [1938, 4225]  &$0.61\pm0.37$ & $\mathbf{0.69}\pm0.36$\\

    Seq4 &[54, 69] &[-8, -7] & [2642, 4074]  &$0.62\pm0.33$ & $\mathbf{0.94}\pm0.04$\\

   Avg &- &- &-   &$0.68\pm0.30$ &$\mathbf{0.83}\pm0.16$\\
   \hline{}
   Times/frame &- &- &-  & 3.0 s & 0.3s \\
   \cline{1-6}
  \end{tabular}
  
  \vspace{1mm}
  \caption{\textbf{Quantitative comparison}. The table shows pan, tilt and focal length ranges in each sequence of the highlights dataset. The most right two columns are IoUs and corresponding standard deviations using different methods. The bottom line shows the average running time per frame for each method. The best performance is highlighted.}
   \label{table:ptz_error}
\end{table*}

In Table \ref{table:ptz_error}, we present mean IoU scores of different methods. The average IoU of our method is significantly higher (0.83 vs. 0.68) than the
CalibMe method. Also, our method is faster (0.3 vs. 3.0 seconds) than CalibMe for two reasons. First, the running time of CalibMe increases linearly with the number of reference images. Because each narrow FOV image covers a small part of the stadium, CalibMe requires a large number of reference images to produce a good result. On the other hand, our pan-tilt forest groups nearby rays in the same leaf node, effectively encoding a large number of reference images. Second, our method requires fewer points (2 vs. 4) to estimate the initial camera parameters than CalibMe. Thus, it needs fewer iterations to find the optimal solution in RANSAC. 

Among four sequences, our method has a relative smaller IoU score in sequence 3. We found that the main reason is that sequence 3 has a larger range of pan angles and some of them are not covered by the training data.

\begin{table}
 \centering 
  \begin{tabular}{l|cc}
    \toprule    
    Games     &CalibMe    &Ours\\ \hline    
  
    BRA vs. MEX  & $0.84\pm 0.24$  & $0.99\pm 0.01$   \\
    BRA vs. NED  & $0.69\pm 0.37$  & $0.98\pm 0.01$   \\
  
   \bottomrule
   
  \end{tabular}
  
  \vspace{1mm}
  \caption{\textbf{World Cup dataset results.}}
   \label{table:world_cup_error}
\end{table}

Table \ref{table:world_cup_error} shows the results on the World Cup dataset. Our method achieves very high IoU scores in this dataset because the images in this dataset cover larger areas than those in the highlights dataset.

\begin{figure}
	\begin{center}
		\includegraphics[width = 1.0\linewidth]{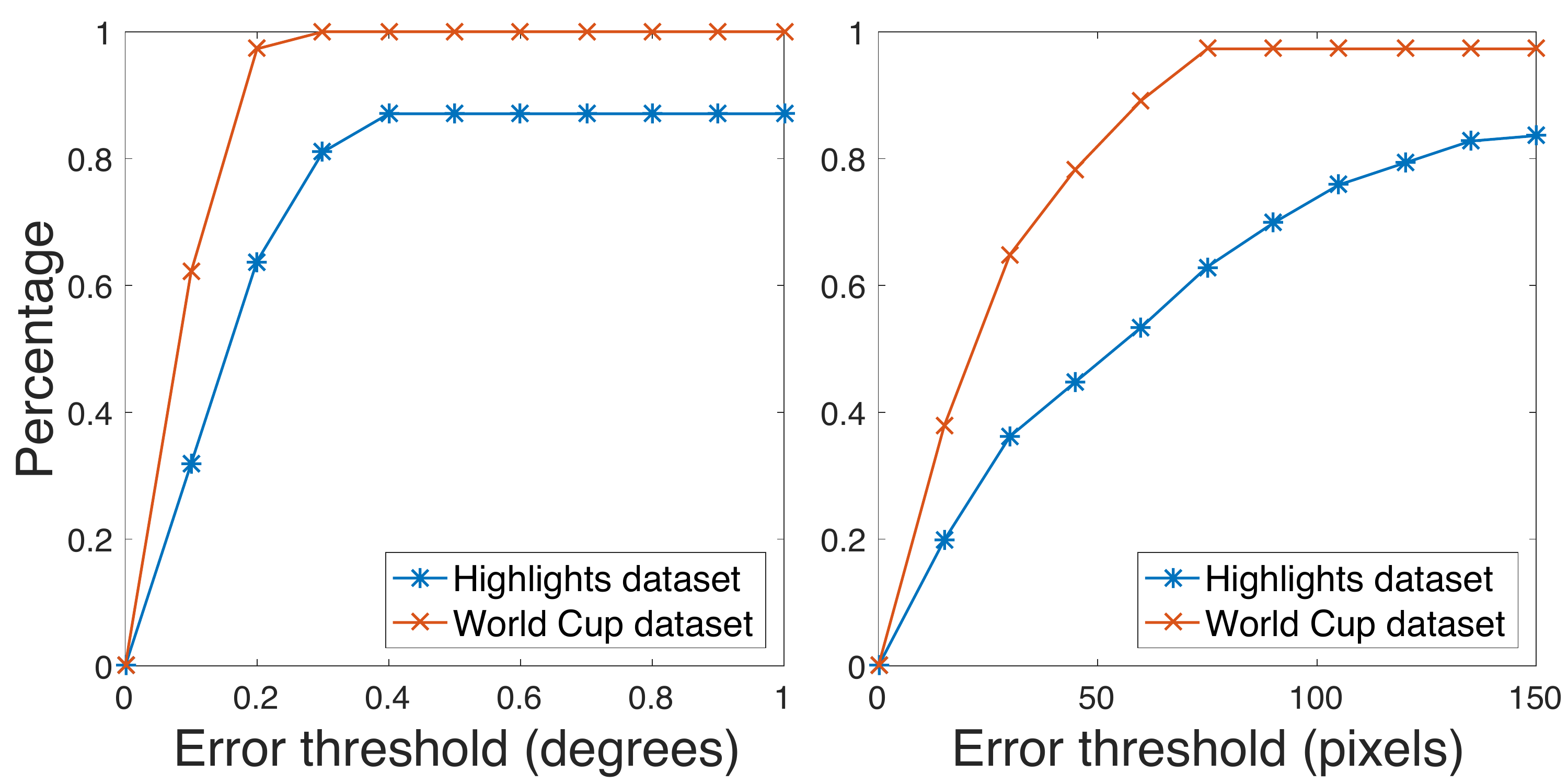}  
	\end{center}
	\vspace{-4mm}
	\caption{\textbf{Cumulative distribution of the rotation and focal length estimation error on two datasets.} Left: rotation errors; right: focal length errors. For the focal length, the relative error of 100 pixels is about 3.2\% of the ground truth.}
	\label{fig:ptz_error_ours}
\end{figure}

Figure \ref{fig:ptz_error_ours} shows the errors of our method measured by rotation error (degrees) and focal length error (pixels), separately. The rotation errors of more than about 85\% of images are less than 1 degree. Our method has small focal length errors in both datasets, with the good performance being especially pronounced in the World Cup dataset. 

Figure \ref{fig:qualitative_result} shows qualitative result of our method. In most cases, our method can calibrate pan-tilt-zoom cameras from very challenging camera angles. 

\vspace{-0.1in}
\paragraph{Further Analysis}

\begin{table}
 \centering
  \begin{tabular}{ l|cc}
    \toprule
   
         & $Inlier_{0.5}$ & Mean IoU  \\ \midrule
    w/o threshold    & 0.09  & 0.53 \\
    w threshold     & 0.33   & 0.83 \\ \hline
    improvement     & +0.24 & +0.30   \\
  \end{tabular}
  \vspace{1mm}
  \caption{\textbf{Influence of using the feature distance threshold in our method}. $Inlier_{0.5}$ means the inlier percentage of pan-tilt predictions that has angular error less than 0.5 degrees. }
   \label{table:rf_vs_ours}
\end{table}

Table \ref{table:rf_vs_ours} shows the influence of with and without the feature distance thresholding in pan/tilt angle prediction. Our method removes outliers using a simple feature distance threshold, resulting in a much higher (0.33 vs. 0.09) percentage of inliers (rotation error less than 0.5 degrees) in pan/tilt angle prediction. The accurate pan/tilt prediction provides a significantly higher (0.83 vs. 0.53) mean IoU score. We also found that the local patch descriptor method is also important for a good performance. For example, if we replace the SIFT descriptor with the SURF descriptor \cite{bay2008speeded}, the mean IoU score drops from 0.83 to 0.74.

\begin{figure}
	\begin{center}
		\includegraphics[width = 0.95\linewidth]{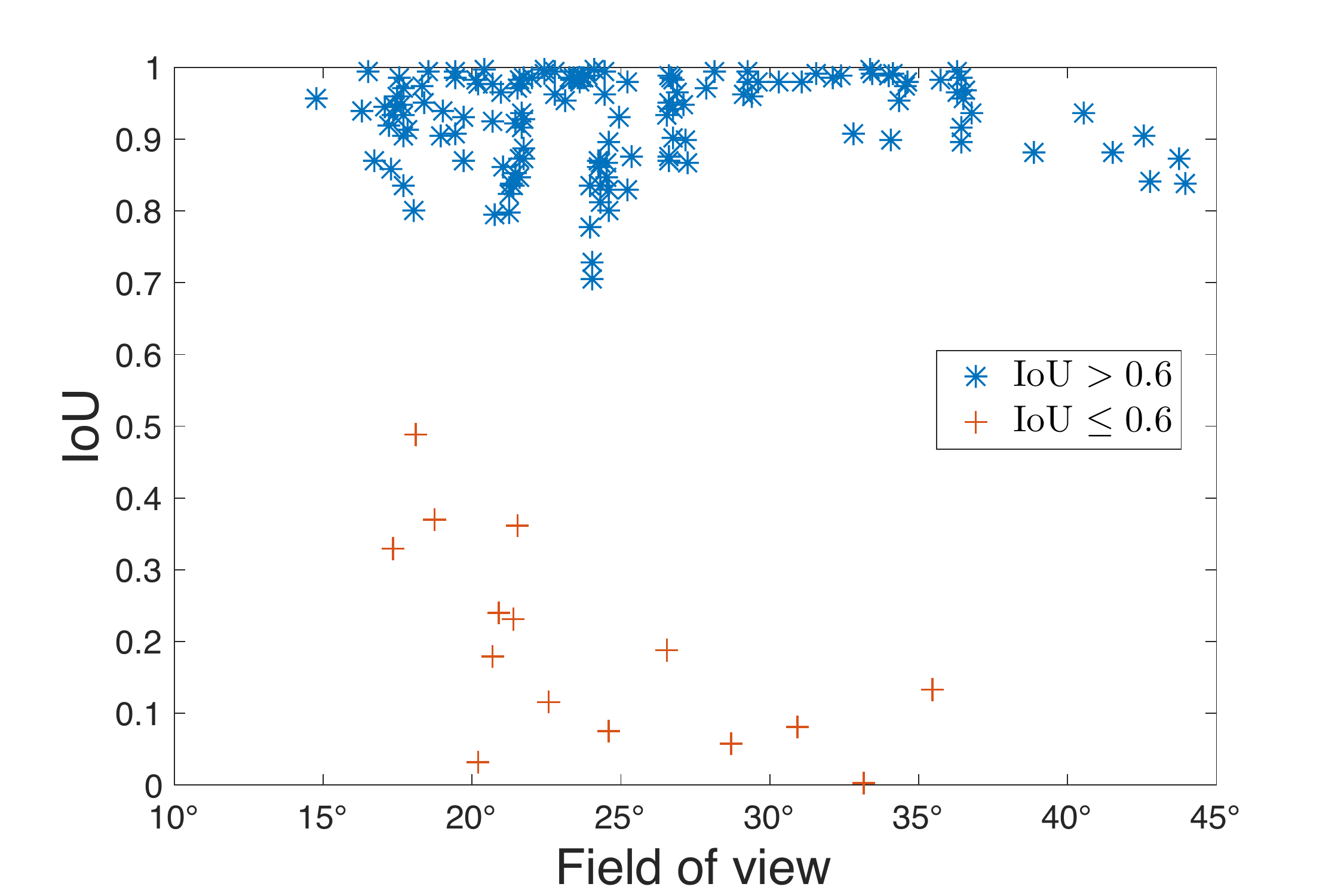}  
	\end{center}
	\vspace{-4mm}
	\caption{\textbf{IoU as a function of the field of view values on two datasets.} }
	\label{fig:iou_vs_fov}
\end{figure}

Figure \ref{fig:iou_vs_fov} shows the IoU measurement as a function of the field of view values on the both datasets of our method. When the FOV values are smaller than about $25^o$, our method has more incorrect estimates (red crosses) whose IoUs are blow a threshold (0.6). When the FOV values are larger than about $40^o$, our method has no incorrect estimates on these two datasets. This result suggests that the narrow field of view still causes incorrect estimates. 

\begin{figure}
	\begin{center}
		\includegraphics[width = 0.95\linewidth]{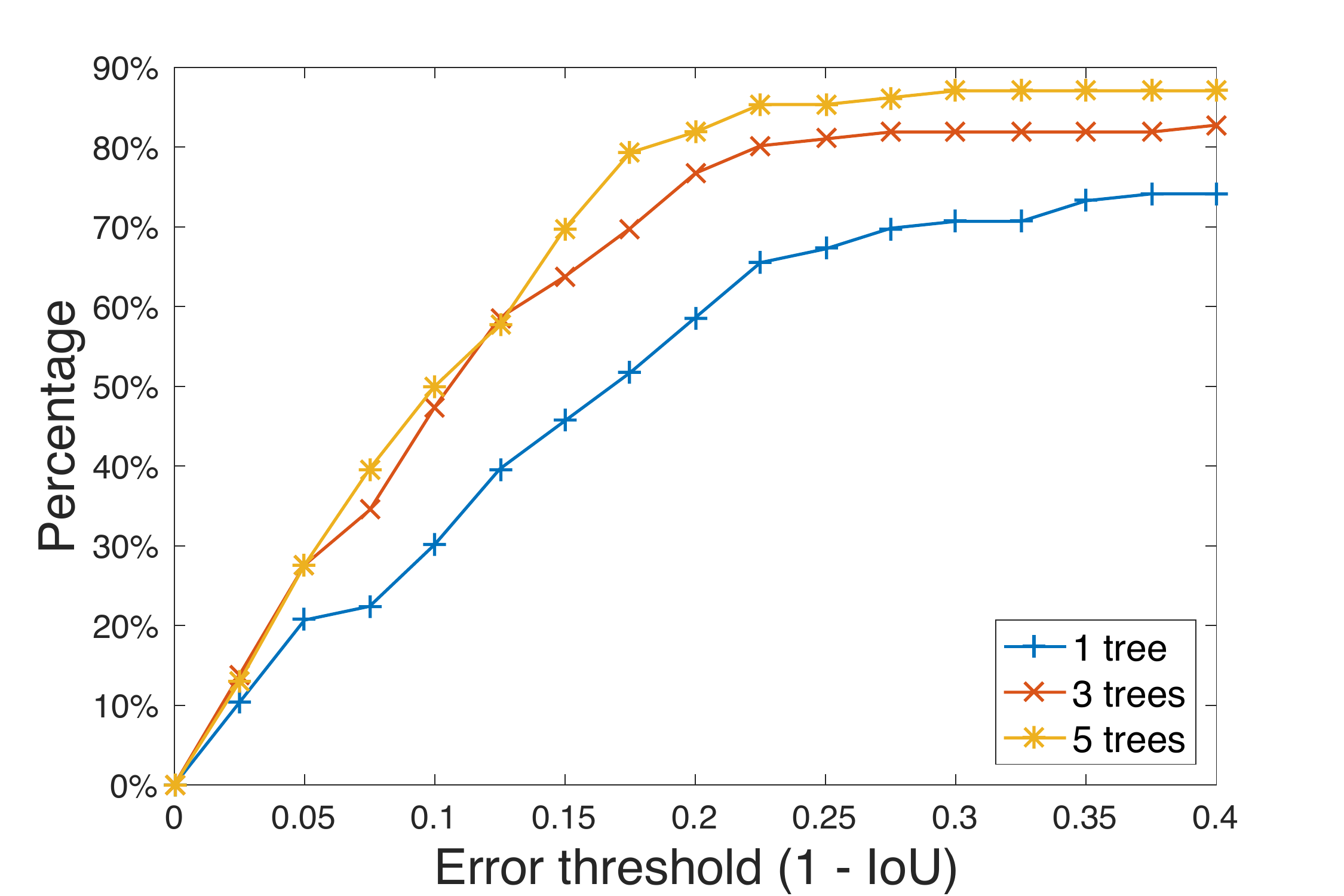}  
	\end{center}
	\vspace{-4mm}
	\caption{\textbf{Impact of tree number.} The figure shows percentage of frames whose calibration errors (1 - IoU) are within a sequence of thresholds. The curve closer to top-left is better.}
	\label{fig:multi_tree_iou_error}
\end{figure}

As an ensemble method, adding more trees also improves the performance at the cost of running time. Figure \ref{fig:multi_tree_iou_error} shows that the performance of our method increases by adding a small number of trees (around 5). 

\vspace{-0.1in}
\paragraph{Implementation Details}
Our approach is implemented with C++ on an Intel 3.0 GHz CPU, 16GB memory Mac system. For the pan-tilt forest, the parameter settings are: tree number is 5, maximum depth is 20. In the test, the computation of SIFT feature costs about 0.2 seconds/frame. The pan/tilt prediction and camera pose optimization costs about 0.1 seconds/frame. In the current implementation, the speed is not optimized. The speed can be improved by using a GPU SIFT implementation and a small number of iterations in RANSAC \etc. The code is available online\footnote{\url{https://github.com/lood339/two_point_calib}}.

\subsection{Limitations}
Conversely, there are several limitations of this work. First, it assumes the camera base parameters are available and fixed. However, these parameters may change from one game to another. In that case, fully automatic methods \cite{wen2016court,homayounfar2017sports} can be used to estimate the camera base parameters before using our method. Second, our camera model does not consider lens distortion which may be significant in some games. One solution is to jointly optimize camera pose and lens distortion \cite{kukelova2015radial} in the camera pose optimization stage. Third, our method uses a real-value feature descriptor which is computationally expensive in running time. One option is to use binary features such as LATCH \cite{levi2016latch}. 

\section{Discussion}
In this work, we proposed a two-point algorithm to efficiently annotate PTZ cameras and a pan-tilt forest method to automatically estimate camera parameters in soccer games. To the best of our knowledge, we are the first to use the two-point method in a random forest framework to calibrate narrow FOV sport cameras. More importantly, our method outperforms the state-of-the-art method in terms of accuracy and speed. 

Currently, we have only evaluated our method on soccer games. However, we expect a similar approach will work for other sports (\eg basketball and ice hockey) as well. In the future, we will also investigate using line and ellipse features \cite{puatruaucean2017joint} in our two-point framework.

\input{appendix}

{\small
\bibliographystyle{ieee}
\bibliography{egbib}
}

\end{document}

%% file: macro.tex
\newcommand{\doubleQuote}[1]{\lq\lq{#1}\rq\rq}

\newcommand{\X}{\textbf{X}}

\newcommand{\x}{\textbf{x}}

\newcommand{\p}{\mathbf{p}}
\newcommand{\bd}{\mathbf{d}}
\newcommand{\br}{\mathbf{r}}
\newcommand{\tK}{\mathtt{K}}

%% file: appendix.tex
\section{Appendix}
\subsection{Focal Length from Two Points}
\label{subseq:app_focal_length}
\begin{equation}
\begin{split}
&f^2 = \\
&\frac{2(d^2ab-c^2)}{2c-d^2(a+b)+\sqrt[]{(d^2(a+b)-2c)^2 - 4(d^2ab-c^2)(d^2-1)}}
\end{split}
\end{equation}
where
\begin{align*}
a &= \x_1^T\x_1 \;\;\;\;\;\; b = \x_2^T\x_2 \\
c &= \x_1^T\x_2 \;\;\;\;\;\; d = \bar{\X}_1^T \bar{\X}_2
\end{align*}
where $\bar{\X} = \frac{\X - \mathbf{C}}{\| \X - \mathbf{C}\|}$, $\X$ are 3D world points, $\x$ are pixel locations and $\mathbf{C}$ is the camera center.

\subsection{Single Point PT Camera Calibration}
\label{subsec:app_single}
Recall our camera model:
\begin{equation}
	\mathtt{P} = \mathtt{K} \mathtt{Q}_{\phi}\mathtt{Q}_{\theta} \mathtt{S} [\mathtt{I} | -\mathbf{C}]	
\end{equation}

There is one 3D point $P_w$ and its projection $p_i$ in the image. We have 
$$ (X, Y, Z) = S(P_w - C)$$ and
$$ (U, V, 1)^T = (K^{-1}p_i)^T.$$
As a result, we have
\begin{equation} 
\begin{bmatrix}
   U \\
   V \\
   1 
   \end{bmatrix}
   = 
	\mathtt{Q}_{\phi}\mathtt{Q}_{\theta} 
     \begin{bmatrix}
	X \\
    Y \\
    Z
    \end{bmatrix}
   = \begin{bmatrix}
	c_p     & 0 & -s_p      \\
    s_t s_p &  c_t & s_t c_p\\
    c_t s_p & -s_t & c_t c_p
	\end{bmatrix} 
    \begin{bmatrix}
	X \\
    Y \\
    Z
    \end{bmatrix}
    \label{equ:pt_proj}
\end{equation}
where $s_p$, $c_p$, $s_t$ and $c_t$ are short for $\sin(pan), \cos(pan), \sin(tilt), \cos(tilt)$. Because the cross product of two sides of \eqref{equ:pt_proj} is a zero vector (\ie cross produt of a vector with itself), we have:
\begin{equation}
\begin{bmatrix}
	A_{pan} & B_{pan}
	\end{bmatrix}
	\begin{bmatrix}
		c_t \\
    	s_t \\
    	1
	\end{bmatrix}
	= \begin{bmatrix}
		0 \\
    	0 
	 \end{bmatrix}
     \label{equ:zero_vector}
\end{equation}
where 
\begin{align*}
A_{pan} &= \begin{bmatrix}
			V X s_p + V Z c_p - Y	&  - VY - X s_p - Z c_p \\
            UX s_p + UZ c_p    & -YU \\
			\end{bmatrix} \\
B_{pan} &= \begin{bmatrix}
				0 \\
                Z s_p - X c_p
			\end{bmatrix}
\end{align*}
From \eqref{equ:zero_vector}, we have
\begin{equation}
c_t =  \frac{ (VY + X s_p + Z c_p) (X c_p - Z s_p)}{det_{Apan}}
\label{equ:tilt_1}
\end{equation}
and
\begin{equation}
s_t = \frac{ (V X s_p + V Z c_p - Y) (X c_p - Z s_p)}{det_{Apan}}
\label{equ:tilt_2}
\end{equation}
where $det_{Apan} = U(Y^2 + (Z c_p + X s_p)^2)$.
Because $c_t^2 + s_t^2 = 1$, we have a quadratic equation of $t_p$ (short for $\tan(pan)$) by
\begin{equation}
a {t_{p}}^2 + b {t_{p}} + c = 0
\label{equ:quad}
\end{equation}
where 
\begin{align*}
a &= (V^2 + 1)Z^2 - U^2(X^2 + Y^2) \\
b &= -2XZ(U^2 + V^2 + 1) \\
c &= (V^2 + 1)X^2 - U^2(Y^2 + Z^2).
\end{align*}
From \eqref{equ:quad}, we can calculate $pan$ angle up to 2 solutions. We can eliminate one of them by setting the valid range to $(-90^o, 90^o)$. The tilt angle can be calculated from \eqref{equ:tilt_1} and \eqref{equ:tilt_2}.